# Follow the Soldiers with Optimized Single – Shot Multibox Detection and Reinforcement Learning

Jumman Hossain
Department of Information Systems
University of Maryland Baltimore County
jumman.hossain@umbc.edu

Maliha Momtaz
Department of Information Systems
University of Maryland Baltimore County
mmomtaz1@umbc.edu

*Abstract*— **Nowadays, autonomous cars are gaining traction due to their numerous potential applications on battlefields and in resolving a variety of other real-world challenges. The main goal of our project is to build an autonomous system using DeepRacer which will follow a specific person (for our project, a soldier) when they will be moving in any direction. Two main components to accomplish this project is an optimized Single – Shot Multibox Detection (SSD) object detection model and a Reinforcement Learning (RL) model. We accomplished the task using SSD Lite instead of SSD and at the end, compared the results among SSD, SSD with Neural Computing Stick (NCS) and SSD Lite. Experimental results show that SSD Lite gives better performance among these three techniques and exhibits a considerable boost in inference speed (~2-3 times) without compromising accuracy.**

*Keywords*—**Autonomous Systems, DeepRacer, Reinforcement Learning, SSD.**

## I. INTRODUCTION

Autonomous vehicles have a lot of importance in real life applications. They are extremely useful in many purposes like in battlefield, mine hunting and imaging through forests etc. The main goal of our project is to build an autonomous system so that it can follow a soldier. This project can be very useful in many aspects. Soldiers can take these autonomous vehicles in the battlefield which will be following them, and it can reduce required number of soldiers in the battle, if those vehicles are equipped with materials needed for battles and can perform war for them. This could save the lives of many soldiers. Besides, these following vehicles can accomplish a very significant task know as Reconnaissance, which is a mission to obtain information by visual observation or other detection methods, about the activities and resources of an enemy or potential enemy, or about the meteorologic, hydrographic, or geographic characteristics of a particular area [1], if they are programmed to do so. Humans and sometimes drones (such as heavily wooded areas) may not be able to accomplish such activities so efficiently by themselves.

In this work, we have optimized the SSD model for faster inference without any loss of accuracy and also enhanced navigation by mapping precise thresholds for action values in order to provide a safe and meaningful action selection. On the AWS DeepRacer device, we employed the OpenVino Optimized SSD object-detection model, which was well-suited for running high-performance inference with the least amount of delay. SSD is much faster compared to two shot RPN (Regional Proposal Network) based approaches. RPN based approaches need two steps to detect objects. In the first step, different regions are proposed and in the second step, multiple objects are detected from these proposed networks. R-CNN is an example of RPN based networks. On the other hand, SSD only need one single step to detect multiple objects. Therefore, SSD is much faster than RPN based approaches. We have used SSD Lite instead of SSD and experimental results show that SSD Lite produced better results.

## II. RELATED WORk

There have been several projects already been developed using AWS DeepRacer device. In the DeepDriver [2] project they mimicked a real-world car that starts and stops at traffic lights and stop signs. The logic for recognizing different colors in traffic signals and detecting stop signs was built by merging several computer vision skills, including OpenCV image processing tools and object recognition machine learning models. Off-road [3] project demonstrated a novel method for autonomously navigating an AWS DeepRacer device across a custom path defined by a series of QR codes serving as checkpoints. To progress along the path, the AWS DeepRacer decodes these QR codes and estimates the next step's direction, and anyone can design their own unique path by sequentially arranging the waypoint codes along the track. The Mapping [4] project created a map using AWS DeepRacer and SLAM (Simultaneous Localization and Mapping), a technique for mapping an environment by predicting a device's present location as it moves across space. In the Follow the Leader (FTL) project [5], it makes use of an object-detection machine learning model to enable the AWS DeepRacer device to recognize and track a human. The FTL allows to extend it by modifying the code to recognize other objects for different use cases with developing custom model, navigation logic, and add-on hardware (optional) to invent new application. In this consideration, we tried to extend it by changing its object detection model with SSD [6] lite and modifying the logic in our use cases for following a soldier whereas it significantly improved the inference performance without sacrificing the accuracy.

## III. SINGLE-SHOT MULTIBOX DETECTION

Deep learning networks are very powerful tools and capable of performing high level image classification tasks. However, it is not wise to compare the ability of humans to machines for such tasks. Humans can localize and classify things very easily while machines are still struggling to perform complex tasks [3].

A few years ago, researchers developed the Region-Convolutional Neural Network (R-CNN) [8] in order to perform object detection, localization and classification. R-CNN is a Region Proposal Network (RPN) based algorithm. An R-CNN is a special type of CNN that is able to locate and detect objects in images. The output is generally a set of bounding boxes that closely match each of the detected objects, as well as a class output for each detected object. Many improvements of R-CNN were proposed later, such as, Fast R-CNN [9], Faster R-CNN [10] etc. Some problems are noticed with these networks. Training the dataset may take too long and training happens in multiple phases. Besides, the network can be very slow at the inference time. In order to overcome these problems, networks like SSD were proposed [6].

Single Shot means that the tasks of object localization and classification are done in a single forward pass of the network. Multibox is the name of a technique for bounding box regression developed by Szegedy et al. [6]. The network is an object detector that also classifies those detected objects (Fig. 1).

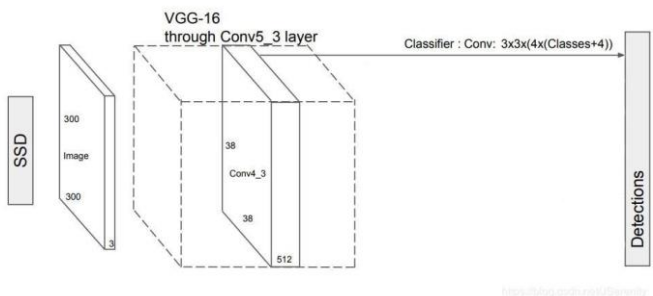

Figure 1: SSD architecture

SSD's architecture builds on the venerable VGG-16 [11] architecture but discards the fully connected layers. The reason VGG-16 was used as the base network is because of its strong performance in high quality image classification tasks and its popularity for problems where transfer learning helps in improving results. Instead of the original VGG fully connected layers, a set of auxiliary convolutional layers (from conv6 onwards) were added, thus enabling to extract features at multiple scales and progressively decrease the size of the input to each subsequent layer.

This network makes multiple predictions containing boundary boxes and confidence scores; therefore, it is called multibox, which is a method for fast class-agnostic bounding box coordinate proposals. It computes both the location and class scores using small convolution filters. SSD uses VGG16 to extract feature maps. Then it detects objects using the Conv4_3 layer. For each cell (also called location), it makes 4 object predictions. Each prediction consists of a boundary box and 21 scores for each class (one extra class for no object), and the highest score is picked as the class for the bounded object. Conv4_3 makes a total of $38 \times 38 \times 4$ predictions: four predictions per cell regardless of the depth of the feature maps. Many predictions contain no object and SSD reserves a class "0" to indicate it has no objects. SSD computes both the location and class scores using small convolution filters. After extracting the feature maps, SSD applies $3 \times 3$ convolution filters for each cell to make predictions. (These filters compute the results just like the regular CNN filters.) Each filter outputs 25 channels: 21 scores for each class plus one boundary box [12].

Generally, SSD uses multiple layers (multi-scale feature maps) to detect objects independently as they improve accuracy significantly. As CNN reduces the spatial dimension gradually, the resolution of the feature maps also decreases. SSD uses lower resolution layers to detect larger scale objects. SSD adds 6 more auxiliary convolution layers after the VGG16. Five of them are added for object detection. In three of those layers, we make 6 predictions instead of 4. In total, SSD makes 8732 predictions using 6 layers.

**SSD vs SSD Lite:** Key important difference of SSD Lite compared to SSD is that the backbone of SSD Lite has only a fraction of the weights of the latter. Therefore, in SSD Lite, the Data Augmentation focuses more on making the model robust to objects of variable sizes than trying to avoid overfitting. Consequently, SSD Lite uses only a subset of the SSD transformations and this way it avoids the over-regularization of the model [13].

## IV. OVERALL

We would like to give some descriptions about the AWS DeepRacer we have worked with. Deep Racer is a fully automated 1 / 18th size racing car driven by RL [14]. It can perform on the global autonomous racing league or compete virtually from anywhere in the world. RL models can be trained, evaluated, and tuned in the online simulator and deployed onto the physical deep racer for a real-world autonomous driving experience (Fig 2.) The car has many different parts. It has an adapter between the power bank and the car. It can be seen as a two-sided USB-C connector. It has USB to USB micro cable, which is used to connect the car to the computer for things such as model transfer. It also has a charger for the car battery, an adapter to charge the compute battery and a power bank which is used to power the computer on the car.

If the hood is removed from the car, many other parts can be seen. It contains an HD camera which is used to take images of the terrain in front of the vehicle. It is used to see where the car is going. Two cameras can be used to set up and train models, if necessary. It has two additional USB ports on two sides, which are left for future expansion. Besides, it has a full-size HDMI port, which helps to enable plugging the car into a monitor, a micro-USB port, which help to enable the setup and model deployment and a micro-USB-C port, which is used to connect to the power bank. It has total 32 gigs of storage, 4 GB RAM and buttons are on and off button, reset button. It also has LEDs to indicate whether the device is booted up or connected to Wi-Fi. The car chassis contains the shocks and 4 wheels for driving. The wheels sometimes look like they are not exactly pointing forward. This is because the car is using Ackermann steering, which is a steering method that turns the wheels at slightly odd angles to prevent slipping when the car goes around the corner.

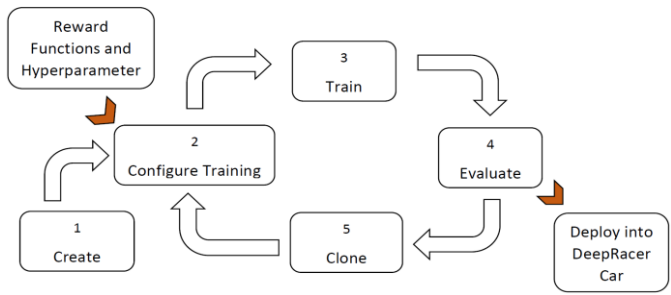

Figure 2: DeepRacer RL model deployment flow

Deep Racer can also contain additional parts. It is possible to integrate two cameras instead of one and when two cameras are added, it works as a stereo camera. Deep Racer with two cameras is called Deep Racer Evo. Using two cameras can be very helpful sometimes because set of inputs become smaller while using only one camera. Stereo cameras are specially important for object avoidance. Deep Racer is also equipped with a LiDAR sensor. The LiDAR sensor allows the DeepRacer to always know its distance to all other objects, such as other DeepRacer's or obstacles on the course. This helps the DeepRacer with obstacle avoidance actions, such as passing another DeepRacer in a Head-to-Head race. Besides, a neural computing stick (NCS) can be connected optionally to the USB slot at the rear end of the car to improve the inference performance.

At first, we planned to solve the problem using our own reinforcement learning model. Reinforcement learning means the car will receive rewards for intended action and punishment for unintended action. Specifically, the agent's (soldiers) position based on the agent (soldiers) action and returns a reward and an updated camera image. The experiences collected in the form of state, action, reward, and new state are used to update the neural network periodically. The updated network models are used to create more experiences. After training, it tries to accomplish the intended action in order to receive rewards.

However, later on we came to know that the DeepRacer device just runs inference on the trained RL models, and the training is provided through AWS console. The open-source device software doesn't consist of the training phase. Hence, it doesn't directly allow to modify the already implemented RL algorithms on the device itself since it just runs inference as part of the AWS-DeepRacer-inference-package which runs with the core application. However, we figured out some possible ways to modify the training to use on our choice of algorithms and environments, including training for real world surroundings. But the problem is - current firmware on the car might struggle with inference for too robust models and a workaround could be to use a remote entity for inference. Considering such issues, we tried to optimize the SSD model for faster inference without sacrificing accuracy and enhance the navigation by mapping precise thresholds for action values in order to provide a safe and meaningful action selection. We used OpenVino Optimized SSD object-detection model that suited to run high-performance inference with minimum latency on the AWS DeepRacer device.

Our system has four major parts: sensor fusion, object detection, navigation, and servo. Sensor fusion node publishes camera images and passes sensor information to object detection node. The object detection node detects bounding boxes and calculate the distance difference between the current person and the new person to be detected. Then object detection node passes the distance difference information to the navigation node. The navigation node then plans actions accordingly and pass information to the servo node, so that it can accomplish the action.

V. METHODOLOGY

At first, the sensor fusion node publishes sensor data from single camera and pass those data to the optimized SSD object detection model. The object detection model detects bounding boxes and calculate normalized delta between current bounding box center and the target center. Normalized delta means the distance difference between the bounding box center of the current person and the bounding box center of the new person to be detected. Then this node passes this normalized delta information to the navigation node and the navigation node plans action based on the data received from the object detection node. The actions which needed to be planned are at what steering angle the car should move to follow the new detected person and at what speed the car should move. The navigation node lets the servo node know about the action plans and map [-1, 1] values to servo calibration. The servo node then sets pwm duty cycle on servo. The same steps happen to the speed to adjust. The follow the soldiers flow diagram is shown below (Fig. 3).

**Inference (Decision):** The inference step is handled by the object detection ROS package, which creates the object detection node that is responsible for collecting sensor data (camera images) from the sensor fusion package and running object detection on the specified object. When an object is detected, the object detection node defines a target center that will serve as a reference for calculating the detection error (delta) everytime the object is detected. This stage involves the node publishing the normalized delta data from the target point as DetectionDeltaMsg data, which contains information about the person or object's location.

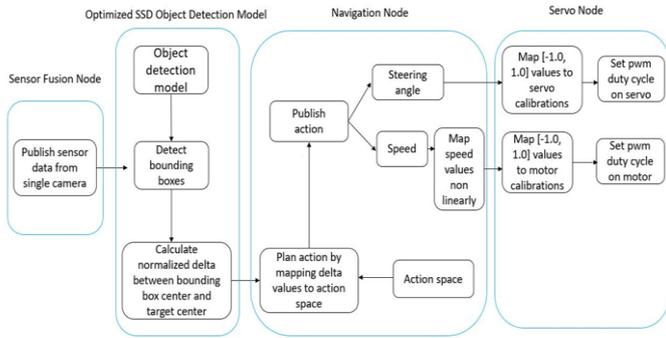

Figure 3: Follow the soldier's flow

The object detection node recognizes an object or a person in each input image. It then gets its bounding box center coordinates and calculates the (x, y) delta between the current position of the detected object and target position (Fig. 4). The DetectionDeltaMsg data is published to the object detection delta topic, which is read by the navigation node. If there is no object detected in an image, the object detection node reports a zero error (delta), indicating that the DeepRacer has already arrived at the target place and does not need to be moved.

**Action (Navigation):** The navigation ROS package creates the navigation node, which determines which action to send based on the normalized detection error (delta) received from the object detection node. The node accounts for the various combinations of expected (x, y) delta values (Fig. 5) using a very simple action space.

There are nine distinct instances (Fig. 6) to consider in relation to the delta x, delta y values. These delta x, delta y parameters specify the distance between the target position and the center of the bounding box in the current image after object detection.

It is important to link particular criteria for the delta x and delta y values to the actions that have been set in order to ensure that actions are selected in a safe and relevant manner.

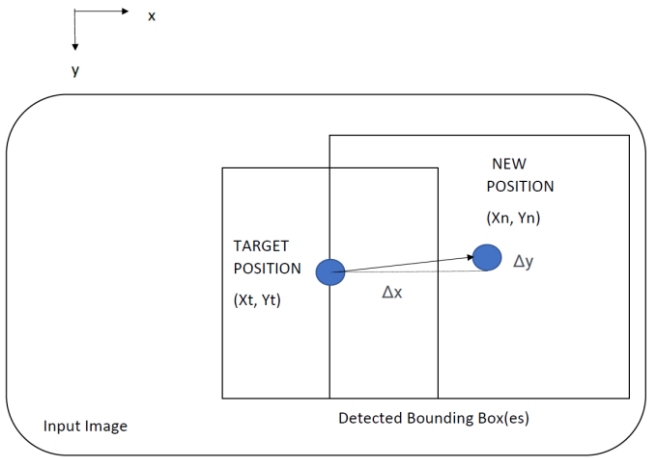

Figure 4: Delta value calculation

The actual delta values used to activate each action in the action space are defined empirically by collecting the delta x, delta y values of the object (person standing in front of camera) at various positions relative to the DeepRacer device's camera. In relation to the object's (person's) position from the camera, we may use these delta x, delta y values to define a safe range of actions. The DeepRacer servo node uses these brackets to map steering and speed parameters.

Based on the brackets of steering and throttle actions, the navigation node plans and publishes an action that the servo node can pick up for every combination of the normalized delta combination in x and y (delta x and delta y). With the help of this pipeline for perception-inference-action on a loop, the DeepRacer detects a soldier (person), plans what action will be required to bring that soldier (person) to the target position, and performs that action for each image on which it infers, thereby achieving the goal of following a soldier (person) in real time.

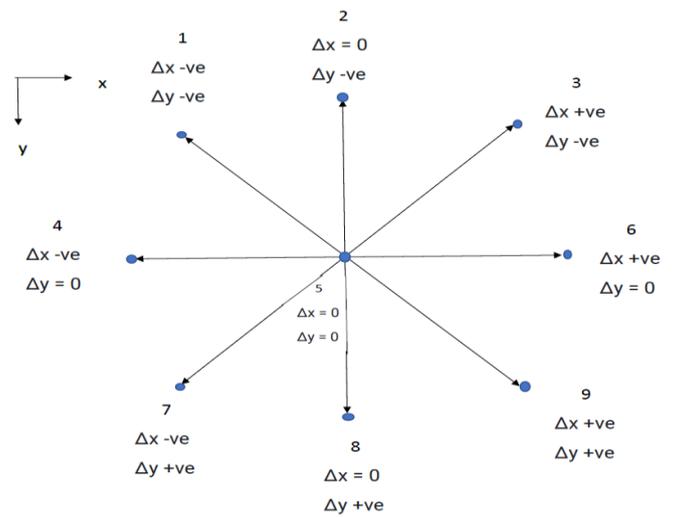

Figure 5: Possible cases for bounding box center movement

| Case | Steering | Throttle |
|------|----------|----------|
| 1 | Left | Forward |
| 2 | NULL | Forward |
| 3 | Right | Forward |
| 4 | Left | NULL |
| 5 | NULL | NULL |
| 6 | Right | NULL |
| 7 | Right | Back |
| 8 | NULL | Back |
| 9 | Left | Back |

Figure 6: Cases to handle in action Space

## VI. EXPERIMENTAL RESULTS

We have implemented the follow the soldier's project using SSD with Neural Computing Stick (NCS) and SSD Lite. Our experimental results show that, SSD with NCS provides better results than SSD and SSD Lite provides the best results among the three. SSD Lite exhibits a considerable boost in inference speed (~2-3 times) without compromising accuracy. The experiment has been conducted while the car was moving in both forward and backward direction. A table containing the execution time for each action (forward or backward) has been included in [Fig. 7], and the time was calculated in milliseconds.

| Direction | SSD | SSD with NCS | SSD Lite |
|---|---|---|---|
| Forward | 0.42902 | 0.22644 | **0.22481** |
| | 0.39288 | 0.22303 | **0.20337** |
| | 0.52259 | 0.23443 | **0.22549** |
| | 0.49419 | 0.19288 | **0.22973** |
| | 0.45433 | 0.19351 | **0.22974** |
| | 0.48836 | 0.25596 | **0.22085** |
| | 0.38706 | 0.21868 | **0.23718** |
| | 0.38466 | 0.19668 | **0.26686** |
| | 0.37103 | 0.19136 | **0.22443** |
| | 0.40832 | 0.18857 | **0.23024** |
| | 0.42056 | 0.21311 | **0.24348** |
| Reverse | 0.45072 | 0.19344 | **0.23044** |
| | 0.38128 | 0.22119 | **0.20888** |
| | 0.46506 | 0.19535 | **0.22706** |
| | 0.39139 | 0.18449 | **0.18941** |
| | 0.49244 | 0.19311 | **0.21787** |
| | 0.38903 | 0.18408 | **0.25114** |
| | 0.44877 | 0.19841 | **0.22736** |
| | 0.44536 | 0.22775 | **0.25937** |
| | 0.49877 | 0.22661 | **0.24518** |
| | 0.46849 | 0.24781 | **0.24683** |
| | 0.47263 | 0.19997 | **0.19531** |
| | 0.42865 | 0.21354 | **0.18401** |

Figure 7: Execution time (ms) for each action

## VII. CONCLUSION AND FUTURE DIRECTION

Currently, the Deep Racer can detect only one soldier (person) at a time and follow that particular soldier (person). Our plan is to extend the model in such a way, so that it can detect multiple soldiers (person), follow a particular one with the help of a marker among those multiple soldiers (person) and update it accordingly. We have been able to detect multiple soldiers but the task of integrating the code on the Deep Racer device is remaining. We would also like to design an object-avoidance model using RL along with following a soldier. (Currently a RL reward function designed where it can avoid box objects in simulation) (Fig. 9).

Apart from the aforementioned two aspects, we will attempt to leverage add-ons such as a depth camera to compute the distance to an item and extend the navigation node concept to utilize the depth aspect to fine-tune the object tracking capacity.

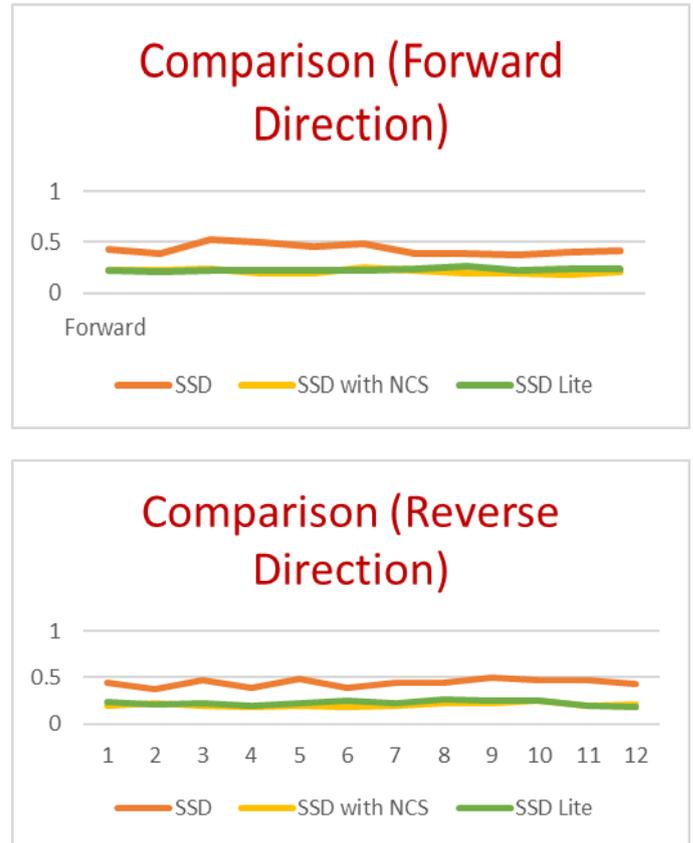

Figure 8: Comparison of SSD, SSD with NCS, and SSD Lite

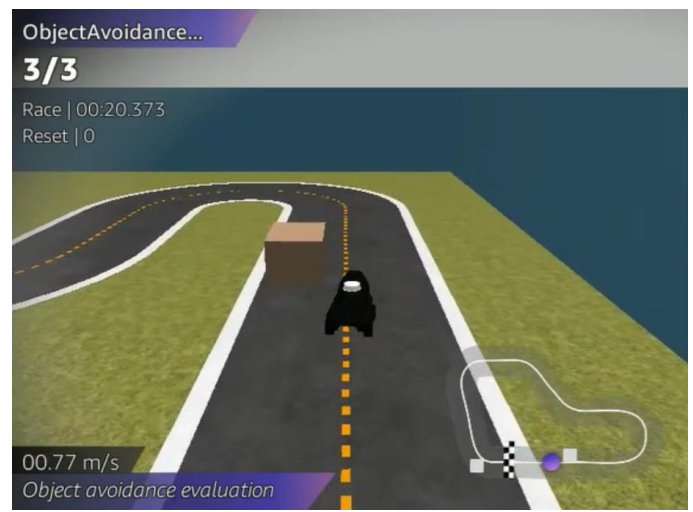

Figure 9: Object avoidance RL model performing in simulation

We would also like to add some interesting ideas, which could be implemented in the future using the Deep Racer device. A security bot can be created which identifies suspicious activities and reports it by moving safely towards the cause, unlike security cameras which can have blind spots. To implement this, we need to change the type of object which needs to be detected instead of a person. Also, the object detection part of the Deep Racer can be programmed in such a way so that it sounds alarms whenever it detects any dangerous substance like gun, knife, smoke etc.

## VIII. ACKNOWLEDGMENT

This research is supported by NSF CNS- 2050999 and U.S. Army Grant No. W911NF2120076.